\documentclass{article}
\usepackage{amsmath}
\usepackage{booktabs} 
\usepackage{caption} 
\usepackage{subcaption} 
\usepackage{graphicx}
\usepackage{pgfplots}
\usepackage[all]{nowidow}
\usepackage[utf8]{inputenc}
\usepackage{tikz}
\usetikzlibrary{er,positioning,bayesnet}
\usepackage{multicol}
\usepackage{algpseudocode,algorithm,algorithmicx}
\usepackage{graphicx}
\usepackage{amssymb}
\usepackage{booktabs}
\usepackage{tabularx}
\usepackage{tabu}
\usepackage{authblk}
\definecolor{blue}{HTML}{1F77B4}
\definecolor{orange}{HTML}{FF7F0E}
\definecolor{green}{HTML}{2CA02C}

\pgfplotsset{compat=1.14}

\begin{document}
\title{Deformable Convolutions and LSTM-based
Flexible Event Frame Fusion Network for
Motion Deblurring}
%
%
\author{Dan Yang, Mehmet Yamac}
%
%
\affil{Huawei Technologies Oy (Finland) Co. Ltd \\
{dan.yang1, mehmet.yamac}@huawei.com\\}
\maketitle              
\begin{abstract}
  Event cameras differ from conventional RGB cameras in that they produce asynchronous 
data sequences. While RGB cameras capture every frame at a fixed rate, 
event cameras only capture changes in the scene, resulting in sparse and 
asynchronous data output. Despite the fact that event data carries useful 
information that can be utilized in motion deblurring of RGB cameras, 
integrating event and image information remains a challenge. 
Recent state-of-the-art CNN-based deblurring solutions produce multiple 2-D event 
frames based on the accumulation of event data over a time period. In most of 
these techniques, however, the number of event frames is fixed and predefined, 
which reduces temporal resolution drastically, particularly for scenarios when 
fast-moving objects are present or when longer exposure times are required. It is 
also important to note that recent modern cameras (e.g., cameras in mobile phones) 
dynamically set the exposure time of the image, which presents an additional problem 
for networks developed for a fixed number of event frames. A Long 
Short-Term Memory (LSTM)-based event feature extraction module has been developed for 
addressing these challenges, which enables us to use a dynamically varying number of 
event frames. Using these modules, we constructed a state-of-the-art deblurring network, 
Deformable Convolutions and LSTM-based Flexible Event Frame Fusion Network (DLEFNet). It 
is particularly useful for scenarios in which exposure times vary depending on factors 
such as lighting conditions or the presence of fast-moving objects in the scene. It has 
been demonstrated through evaluation results that the proposed method can outperform the 
existing state-of-the-art networks for deblurring task in synthetic and real-world data 
sets.
\end{abstract}

\section{Introduction}
\label{sec:intro}
Typically, blur is a common distortion that can be attributed to a variety of factors, including motion, camera shake, depth variations in the scene, and others. In earlier approaches, blurring kernels were applied globally to sharp images or applied locally to sharp images \cite{classic-1}. It is, however, a challenging ill-posed problem to estimate both blur kernels and sharp images simultaneously, which may require iterative computation. With the emergence of neural networks, most recent approaches based on convolutional neural networks (CNNs) have demonstrated significant improvements in reconstruction accuracy and computation efficiency \cite{GoPro, SRN, MPRNET, BANet, MIMO}. Typically, they do not require kernel estimation and train directly to map blurry images to sharp images using data sets that are more realistic, such as the GoPro \cite{GoPro} or REDS \cite{REDS} datasets. Although single-image deblurring networks have made significant advances \cite{BANet, MIMO, Nafnet, restormer}, their performance is still inadequate in challenging situations, particularly when dealing with fast-moving objects.

This study aims to improve the quality of blurry images captured by standard cameras by leveraging the capabilities of an event camera. The bio-inspired sensors used in event cameras are capable of detecting changes in the brightness of pixels at the microsecond level, thus encoding spatiotemporal information of rapidly changing scenes at the pixel level \cite{eventCamera1, eventCamera2, eventCamera3}. Deblurring tasks, especially for images of dynamic scenes with moving objects, may benefit greatly from such information \cite{BHA, LEBMD, MADANet, EFNet}.

A number of advantages are offered by event cameras over traditional cameras, including a higher dynamic range (exceeding 120dB, as opposed to 60dB for traditional cameras), higher temporal resolution, and lower power consumption \cite{eventsurvey}. Nevertheless, event cameras differ from conventional RGB cameras in their asynchronous output of data sequences, which makes applying conventional computer vision algorithms to event data difficult. As compared to traditional cameras, event cameras record only changes in the scene, leading to a sparse and asynchronous output of data. While the use of event data for motion deblurring of RGB cameras is valuable, the integration of event and image data remains a challenge. Based on the assumption that blurry images can be represented as integrals of sharp latent images, \cite{BHA} developed a model for deblurring intensity images by utilizing a double integral technique. Such model-based algorithms are iterative and computationally demanding, which makes them incompatible with portable devices such as smartphones. The algorithms are also limited to grayscale images, so adapting them to RGB images remains challenging.

Researchers have recently focused on integrating event data over a time period in order to create multi-frame event data that is more compatible with computer vision algorithms developed for RGB frame processing. This approach enables the use of multi-frame event data alongside RGB frames for tasks such as deblurring \cite{MADANet, EFNet}. The current literature, however, fixes the number of event frames by assuming a fixed exposure time for all the images, leading to a reduced temporal resolution when dealing with fast-moving objects or when a longer exposure time is necessary. As an example, in mobile cameras, the exposure time varies dynamically, which poses a particular challenge.

The purpose of this study is to develop a neural network structure for event-data-assisted deblurring that can accommodate dynamic changes in the number of event frames being constructed. This network is suitable for two realistic scenarios or a combination of them: (i) Modern cameras, including mobile cameras, are capable of setting the exposure time automatically based on factors such as lighting conditions, ISO, aperture, and shutter speed when capturing images. (ii) If one desires to use a fixed exposure time for an RGB camera, integrating event data using uniformly divided time interval bins to produce event frames is not optimal. In order to achieve better results, smaller bins (i.e., having more event frames) can be used when more event sensors are triggered, such as when cameras are moving faster or when objects are moving faster.

First, we have developed a feature extraction module based on Long Short-Term Memory (LSTM) that allows us to use a dynamic number of 
event frames during feature extraction. Then, we have used deformable CNNs in the LSTM module in order to increase the functionality 
of feature extraction. Afterward, the extracted features are combined with the encoded features of RGB frames in multiple scales to construct a state-of-the-art deblurring network, DLEFNet.  
In addition to its ability to handle fast-moving objects and varying exposure times, 
our model proves particularly valuable in real-world scenarios where the scene properties is unknown in advance. 
The proposed solution has achieved new state-of-the-art levels of image reconstruction accuracy for both the GoPro dataset \cite{GoPro} and the 
recently released real event data deblurring dataset, REBlur \cite{EFNet}.

\section{Related Works}
An event camera detects changes in intensity and records the pixel location, change time, and polarity of change. As a result, event cameras obtain sparse and asynchronous information with high temporal resolution, which is significantly different from conventional standard image representation. This makes it challenging to use event data with conventional computer vision algorithms.

One approach is to consider a standard camera coupled with an event camera. In this approach, the blurry image intensity can be regarded as the integral over sharp latent images changing with time, and each small change between any so-called sharp images can be calculated over the sum of the event stream. This assumption led to the proposal of a model for deblurring intensity images using a double integral technique \cite{BHA}.
After that, a deep unfolding neural network has been proposed to model deblurring as a Maximum-a-Posteriori (MAP) problem and solve it \cite{LEBMD}. 

Recent works have shown that event streams can be converted to event frames by simple accumulation for various computer vision tasks such as image and video deblurring \cite{MADANet, EFNet}, depth estimation \cite{depthevent}, optical flow estimation \cite{opticalflow}, and feature tracking \cite{tracking}. Although neural network-based solutions like MADANet and EFNet \cite{MADANet, EFNet} that feed the network with blurry images and event frames obtained through simple uniform quantization with a fixed number of bins achieve state-of-the-art deblurring results, there is still significant room for improvement, particularly in event feature extraction and/or fusion steps. This is because these methods consider a fixed and generally small number of uniformly split bins (e.g., 5, 6, or 7), which fails to utilize the full temporal information that event data provides.

In this work, we propose a novel event-data-assisted deblurring neural network structure that can adapt to the changing number of event frames between captures. This approach offers two significant benefits: (a) it makes the algorithm more suitable for modern cameras whose exposure time dynamically changes, and (b) it allows for smaller interval bins (more event frames) than the literature, which enables the network to utilize more temporal information provided by event data. 

We have developed a novel approach for deblurring networks that leverages a feature extraction module based on Long Short-Term Memory (LSTM) capable of utilizing a dynamic number of event frames during feature extraction. Furthermore, we have incorporated deformable convolutional neural networks (CNNs) within the LSTM module to enhance the functionality of feature extraction. The extracted features are then combined with the encoded features of RGB frames across multiple scales to create a state-of-the-art deblurring network. Our model demonstrates superior performance in scenarios where the exposure time is unknown in advance, making it particularly valuable. Additionally, the model is capable of handling fast-moving objects and varying exposure times.

In a recent paper \cite{vitoria2023event}, convolutional LSTMs and deformable convolutions are utilized as well, but separately. The former is used to extract event features, whereas the latter is used to combine image features with event features. Their aim is to create local motion-aware convolution. However, they still consider a fixed number of event frames (i.e., N=5). The study \cite{kim2022event} might be more relevant to our target task. The study \cite{kim2022event} attempted to handle dynamically changing video deblurring exposure times. However, they still assumed a fixed shutter period, divided into an unknown exposure time and an unknown readout time. Essentially, they considered a predefined number of frames, $N=m+n$, and used an event feature selective module to select features from the first $m$ frames when $m$ was unknown. In contrast, our study focuses on single image deblurring with an unknown exposure time and therefore an unknown number of event frames.

\section{Proposed Solution}
\subsection{Event Data Pre-processing}

In contrast to traditional cameras, newly emerging event cameras do not have a predetermined exposure time. In contrast, they detect changes in brightness asynchronously at the microsecond level within a scene. A sequence of events, denoted by $\left(x,y,t,\delta\right)$, is output by event cameras instead of capturing the intensity at pixel $\left(x,y\right)$ at time $t$. The time of occurrence of the event is represented by $t$, and the polarity of the event is represented by $\delta$:
\[
  \delta  = \begin{cases}
      +1, & g \geq  thr \\
      -1 & g \leq -thr \\
      0 & \text{else} \\
  \end{cases}
\]
where $thr$ is the threshold determined by the the user, $I_{x,y}$ is the intensity level at pixel $\left(x,y\right)$ and the proportional change calculated logarithmically, i.e.,  $g =\log \left ( \frac{I_{x,y}\left ( t \right )}{I_{x,y}\left ( t_{ref}
\right )} \right ) $, when $t_{ref}$ is the previous event occurrence time.

For the purpose of processing these events, instead of dividing the event stream into several chunks during the standard and fixed exposure time, $T$, as is done in literature \cite{MADANet, EFNet}, we consider that $T$ is neither fixed nor known. We accumulate event data over very short intervals of time, $\Delta T$ (e.g., 1 or 2 milliseconds). A ternary 2-D frame is then constructed from the accumulated data, i.e. 
\begin{equation}
  e_{x,y}^i = Q\left(\int_{\left(i-1\right) \Delta T}^{i \Delta T} \varepsilon_{x,y}\left(t\right)dt\right) ~~ \text{for} ~~ i = 1, 2, ..
\end{equation}
where $Q(h) = \text{sgn}(h)$ and $\varepsilon_{x,y}\left(t\right)$ corresponds to an individual event occurring at time $t$ and at pixel location $(x,y)$. Unlike similar event representation schemes \cite{depthevent, opticalflow, tracking}, our event frames have ternary data, $e_{x,y}^i \in \left\{1,0,-1\right\}$. As the exposure time, $T$, is unknown in advance, the number of event frames varies from image to image, i.e., $n=\frac{T}{\Delta T}$.

\subsection{Deformable Convolutional LSTM for Event Feature Extraction}
Let $\mathbf{B} \in \mathbb{R}^{3 \times H \times W }$ be the blurry RGB image. The Dynamic Event Frame Feature Extraction (DEFFE)
module uses the dynamically changing number of event frames and the blurry image in order to get $C$ number of event feature maps using
Bidrectional Deformable Convolution Long Short-Short Term Memory (BiDefConvLSTM) cells in recurrent manner. 

The Deformable Convolution LSTM (DefConvLSTM) cell used in this module is formulated as follows: 
\begin{align}
 i_t =& \sigma \left ( W_{Ei}  \otimes \mathbf{e^t} +W_{Hi} \otimes \mathbf{H_{t-1}} + W_{Bi} \otimes \mathbf{B}  + b_i \right ),\\
 f_t =& \sigma \left ( W_{Ef}  \otimes \mathbf{e^t} +W_{Hf} \otimes \mathbf{H_{t-1}} + W_{Bf} \otimes \mathbf{B} + b_f \right ),\\
 o_t =& \sigma \left ( W_{Eo}  \otimes \mathbf{e^t} +W_{Ho} \otimes \mathbf{H_{t-1}} + W_{Bo} \otimes \mathbf{B} + b_o \right ),\\
 c_t =& \sigma \left ( W_{Ec}  \otimes \mathbf{e^t} +W_{Hc} \otimes \mathbf{H_{t-1}} + W_{Bc} \otimes \mathbf{B} + b_c \right ),\\
 C_t =& f_t \odot tanh \left ( c_t \right ) \\
 H_t =& o_t \odot tanh \left ( C_t \right ) \\
 w_t =& ha \left ( \textbf{H}_t \right ) \\ 
 \textbf{Fe} =& \mathbf{Fe} + w_t\mathbf{H}_t, 
\end{align}
where $\otimes$ denotes the deformable convolutional operator, $\odot$ denotes the
Hadamard product, $\sigma$ denotes the Sigmoid operation, $i_t$, $f_t$, $o_t$ and
$c_t$ denotes input, forget, output, and cell gates, $W_{E*}, W_{H*}, W_{B*}$ are
deformable convolutional kernels, $b_*$ are bias terms, $w_t$ is hidden layer attention weight, 
$\mathbf{Fe }\in \mathbb{R}^{C \times H \times W}$ is $C$-channel
extracted feature map and $ha(.)$ is hidden layer attention operation. The hidden layer attention
operation is simply channel-wise squeeze operation followed by MLP layers and a sigmoid, i.e.,
for a feature map $\mathbf{X }\in \mathbb{R}^{C \times H \times W}$, it is defined as
\begin{equation}
 ha (\mathbf{X}) = \sigma \left ( MLP \left ( GAP \left ( \mathbf{X} \right ) \right ) \right ) \in \mathbb{R}^{1 \times 1 \times 1}. \label{MLP}
\end{equation}
The proposed LSTM module works on Bidrectional manner in a way that as forfard
process the above mentioned LSTM cells takes firt half of the event frames in 
forward manner, i.e., $1, 2, ..., n/2$. On the other hand, as bacward process the 
cells takes the second half in reverse order, i.e., $n, n-1,..., n/2$.

\begin{figure}[ht]
\centering
\includegraphics[width=0.99\linewidth]{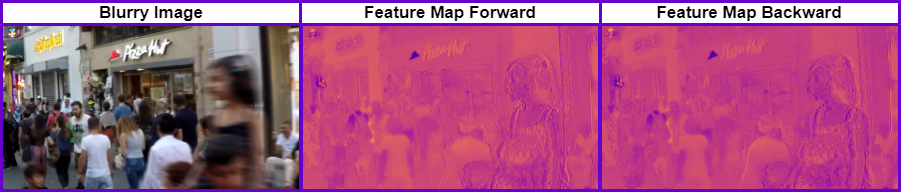}
\caption{An example pair of forward and backward feature maps.}
\label{fig:resultsReblur}
\end{figure}

\subsection{The overall network}
DLEFNet consists of two main parts: the first is the DEFFE Module, which was explained above, and the other is the main body deblurring module, denoted by $\mathcal{M}$. The deblurring module is an adaptation of the Frequency Domain Multiple Input Multiple Output Unet (FMIMO-UNet) structure and consists of an encoder, $\mathcal{M_E}$, and a decoder, $\mathcal{M_D}$. Both the encoder and decoder have a multilevel structure, with three levels.

Let $\mathcal{M_E}_L$ be the $L^{th}$ level encoder block, which takes as input $\mathbf{F{in}}^L \in \mathbf{R} ^{c_i \times h_i \times w_i }$ and $\mathbf{Fe_{in}}^L \in \mathbf{R} ^{c_i \times h_i \times w_i }$, and produces as output $\mathbf{F_{out}}^L \in \mathbf{R} ^{c_o \times h_o \times w_o }$. Specifically,
\begin{equation}
\mathbf{F_{out}^L} \leftarrow \mathcal{M_E}_L \left ( \mathbf{F{in}}^L, \mathbf{Fe_{in}}^L \right ),
\end{equation}
where $\mathbf{F_{in}}^L$ is the input image feature map with $c_i$ channels, and $\mathbf{Fe_{in}}^L$ is the event feature map for the $L^{th}$ level. 
For the first level, the encoder module takes the 3-channel RGB blurry image, $\mathbf{F_{in}}^L= \mathbf{B}$. 
A convolutional layer with a stride of 2 is applied to $\mathbf{F_{out}^L}$ to obtain $\mathbf{F_{in}}^{L+1}$, which doubles the channel dimension and 
reduces the height and width by a scale factor of 2. 

\begin{figure*}[ht]
\centering
\includegraphics[width=0.95\linewidth]{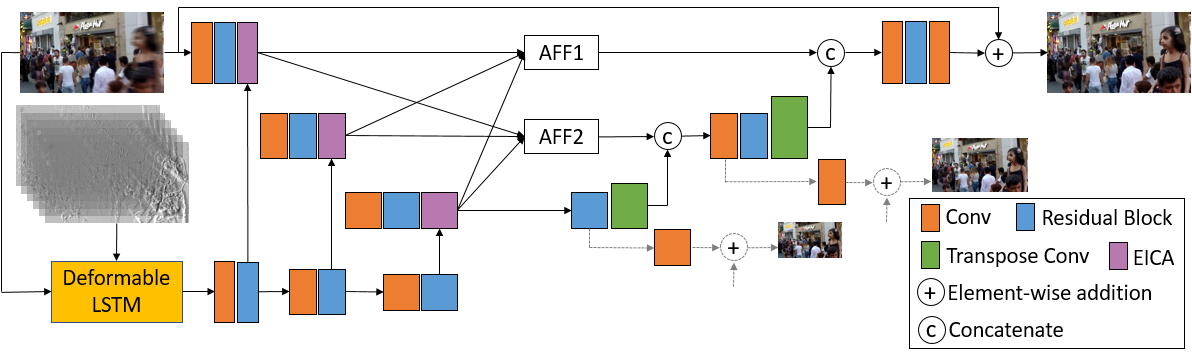}
\caption{The pipeline of the proposed DLEFNet}
\label{fig:model-architect}
\end{figure*}

Each $\mathcal{M_E}_L$ first processes the image feature maps using frequency domain residual blocks, 
as proposed in \cite{DeepRFT}. 
To fuse the event feature maps at each level, the Event-Image Crossmodal Attention (EICA) fusion module, 
borrowed from \cite{EFNet}, is used. For the decoder, $\mathcal{M_D}$, the features are processed by 
frequency domain residual blocks, and multi-scale feature maps are fused using asymmetric feature 
fusion (AFF) modules, as done in MIMO-Unet \cite{MIMO}. The event features from DEFFE moudule is also passed through an $L$ level event encoder block, 
$\mathcal{M_E}^E$ before fusing them with image feature maps. The overall structure of the network can be 
found in Figure \ref{fig:model-architect}.

\section{Network Details and Training}

In this study, we have designed our U-Net architecture with a symmetrical structure. We set the number of output channels to 32 for the first level, 
and then increased it to 64 and 128 in the following layers of the encoder. For the decoder, the number of channels was set to 128, 64, and 32, respectively.
We used $3 \times 3$ kernels for all the convolutional layers. The number of event feature maps was set to 32, which were obtained by considering 16 
forward and 16 backward directions. We employed 8 residual blocks at each level. To compute the loss, we utilized a hybrid loss 
function comprising  $\mathcal{L}_{1}$ ($\ell_1$-norm metric) and a SSIM-based loss $\mathcal{L}_{s}$, i.e., $\mathcal{L} = \mathcal{L}_{1} +  0.5\mathcal{L}_{s} $. We applied a 
cosine scheduler that decreased the learning rate from 0.001 to 1e-7. In MLP block given in Equation \eqref{MLP}, 32 channel was map to 2, then a ReLu activation function
was used before last layer mapping 2 dimensional scalar value defining the which hidden layers are more relevant. 




\section{Experimental Evaluation}
\begin{table} \scriptsize
\centering
\caption{Deblurring results on GoPro dataset}
\label{tab:results_gopro}
\begin{tabular}{lrrr} 
\toprule
Method        & PSNR                        & SSIM                      & Params                         \\ 
\midrule
BHA~\cite{pan2019bringing}        & \multicolumn{1}{l}{~~29.06} & \multicolumn{1}{l}{0.943} & \multicolumn{1}{l}{~ ~ ~ N/A}  \\
DeepDeblur~\cite{nah2017deep}   & 29.23                       & 0.916                     & 11.7M                          \\
SVDN~\cite{yuan2020efficient}         & \multicolumn{1}{l}{~~29.81} & \multicolumn{1}{l}{0.937} & \multicolumn{1}{l}{~ ~ ~N/A}   \\
SRN~\cite{tao2018scale}          & 30.26                       & 0.934                     & 6.8M                           \\
DGN~\cite{DGN}           & \multicolumn{1}{l}{~ 30.49} & \multicolumn{1}{l}{0.938} & \multicolumn{1}{l}{11.32M}     \\
PSS-NSC~\cite{gao2019dynamic}      & 30.92                       & 0.942                     & 2.8M                           \\
MT-RNN~\cite{park2020multi}       & 31.15                       & 0.945                     & 2.6M                           \\
DMPHN~\cite{zhang2019deep}      & \multicolumn{1}{l}{~ 31.20} & \multicolumn{1}{l}{0.945} & \multicolumn{1}{l}{~~21.7M}    \\
RADN~\cite{purohit2020region}         & 31.76                       & 0.953                     & N/A                            \\
LEBMD~\cite{jiang2020learning}      & \multicolumn{1}{l}{~ 31.79} & \multicolumn{1}{l}{0.949} & \multicolumn{1}{l}{~ ~ ~ N/A}  \\
PVDNet~\cite{PVDNet}        & \multicolumn{1}{l}{~ 31.98} & \multicolumn{1}{l}{0.928} & \multicolumn{1}{l}{~ 23.4M}    \\
SAPHN~\cite{suin2020spatially}       & \multicolumn{1}{l}{~ 32.02} & \multicolumn{1}{l}{0.953} & \multicolumn{1}{l}{~ ~ ~~N/A}  \\
GSTA~\cite{GSTA}          & \multicolumn{1}{l}{~ 32.10} & \multicolumn{1}{l}{0.960} & \multicolumn{1}{l}{~ ~ ~~N/A}  \\
MBRNN~\cite{MBRNN}         & \multicolumn{1}{l}{~ 32.16} & \multicolumn{1}{l}{0.953} & \multicolumn{1}{l}{~ 5.42M}    \\
BANET ~\cite{tsai2021banet}       & 32.44                       & 0.957                     & 85.6M                          \\
MPRNET~\cite{zamir2021multi}       & 32.66                       & 0.959                     & 20.1M                          \\
MIMO-UNet++~\cite{cho2021rethinking} & 32.68                       & 0.959                     & 16.1M                          \\
HINet~\cite{chen2021hinet}         & \multicolumn{1}{l}{~ 32.71} & \multicolumn{1}{l}{0.959} & \multicolumn{1}{l}{~ 88.6M}    \\
  Restormer~\cite{restormer}  & 32.92                       & 0.961                     & 26.1                            \\
ERDN~\cite{ERDN}         & 32.99                       & 0.935                     & N/A                            \\ 
  NAFNet~\cite{Nafnet}                   &33.71                        &0.967                      &67.9                            \\
{MADANET+}  ~\cite{MADANet}   & {33.84}                       & {0.964}                     & 16.9M                          \\
  Vitoria et al.~\cite{vitoria2022event}  & 34.33&0.944 &N/A\\
  EFNet~\cite{EFNet}& 35.46&0.972 &8.5M\\
  \midrule
  DLEFNet (ours) & 35.61& 0.973 &12M\\ 
\bottomrule
\end{tabular}
\end{table}
\subsection{GoPro Dataset}

\begin{figure*}[ht]
\centering
\includegraphics[width=0.95\linewidth]{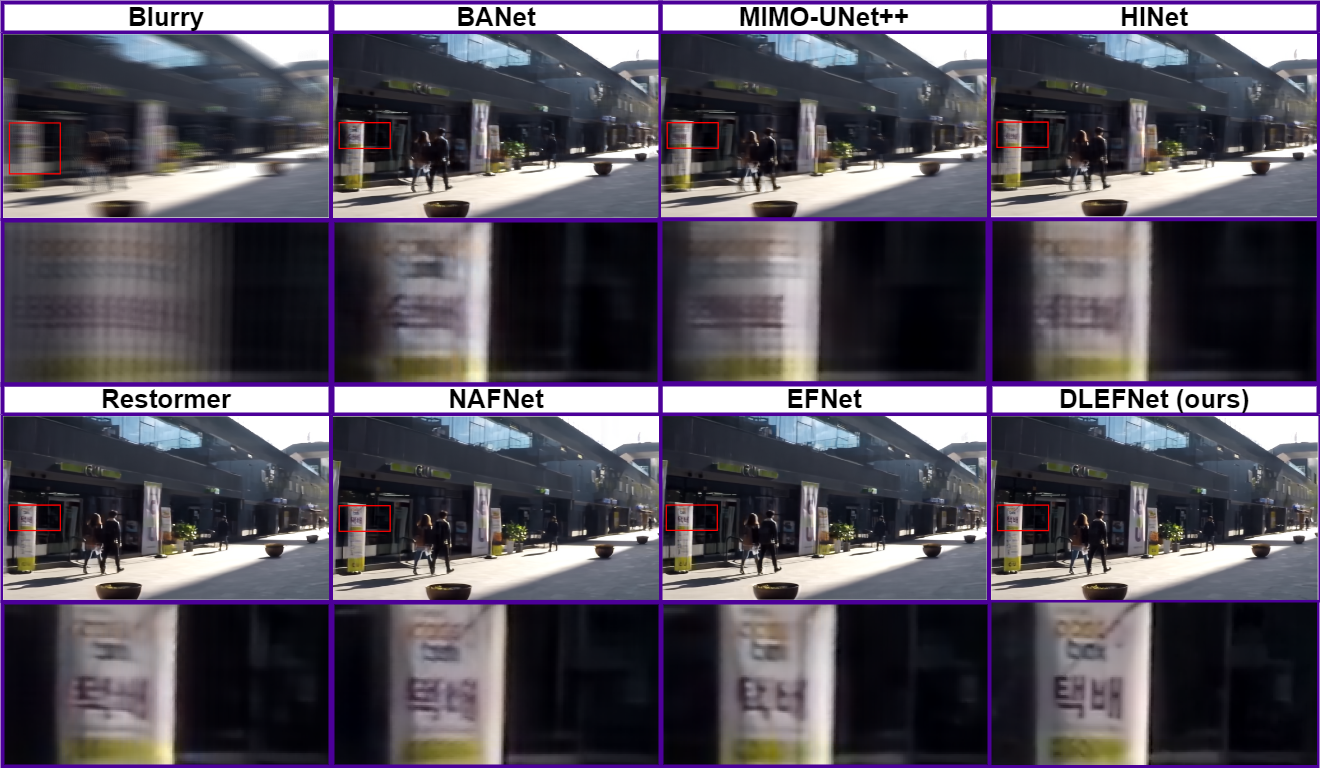}
\caption{Visual comparison of the state-of-the-art algorithms on GoPro dataset}
\label{fig:resultsGoPro}
\end{figure*}

We adhere to the recommended protocol for training and testing by averaging a varying number of nearby frames (ranging from 7 to 13) to generate the blurry image. In that manner, the synthetic events corresponding to the averaged frames are accumulated into varying number of  event frames ranging from 7 to 13. 
ESIM simulator~\cite{rebecq2018esim} was used to generate the corresponding events. We use a total of 2103 samples for training, 
consisting of blurry image, event frames stack, and ground-truth frames, and 1111 samples for testing. 
Table \ref{tab:results_gopro} compares the performance of DLEFNet with other state-of-the-art deblurring methods. 
BHA \cite{BHA}, LEBMD \cite{LEBMD}, ERDN, Vitoria et al. \cite{vitoria2023event}, and EFNet \cite{EFNet} are event-aided solutions, 
while DGN \cite{DGN} provides depth-aware deblurring. MBRNN \cite{MBRNN}, GSTA \cite{GSTA}, and PVDNet \cite{PVDNet} are video-based (multi-frame) methods. 
The proposed solution significantly outperforms the SoTa deblurring techniques as can be seen in the Table \ref{tab:results_gopro}. Visual comparison
of the algorithms are given in Figure \ref{fig:resultsGoPro}.

\subsection{REBlur Dataset}
\begin{figure}[ht]
\centering
\includegraphics[width=0.95\linewidth]{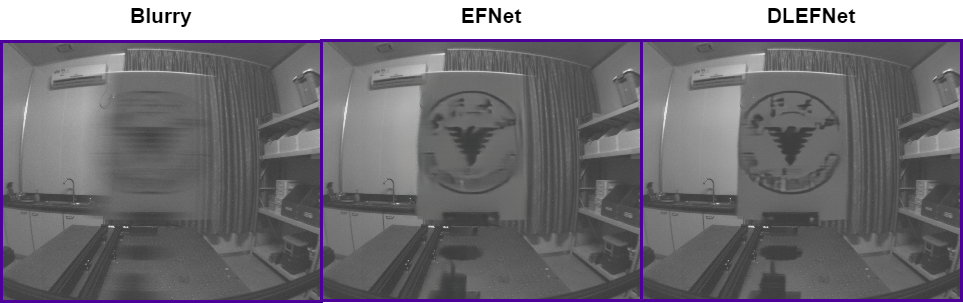}
\caption{Visual comparison of the state-of-the-art algorithms on REBlur dataset}
\label{fig:resultsReblur}
\end{figure}

The Real Event Blur (REBlur) dataset is a recently proposed collection of event data and real blurry images \cite{EFNet}. REBlur encompasses 12 
different types of linear and nonlinear motions for 3 distinct moving patterns and for the camera itself. It includes 36 sequences and 1469 pairs of 
blurry-sharp images with associated events. Of these pairs, 486 are used for training, and 983 for testing. As done in \cite{EFNet}, 
we fine-tuned our network trained on the GoPro dataset with the training portion of the REBlur dataset. 
The evaluation was then conducted on the test set. The performance comparison of algorithms is presented in Table \ref{tab:results_Reblur}. The results of 
the algorithms with a "+" are from \cite{EFNet}, in which the blurry images and event frames produced in \cite{EFNet} are fed and trained together. 
A visual comparison of EFNet and DLEFNet is provided in Figure \ref{fig:resultsReblur}. Our proposed solution outperforms the state-of-the-art network EFNet both quantitatively and visually.
\begin{table}[]
  \centering
\caption{Deblurring results on REBlur dataset}
\label{tab:results_Reblur}
  \begin{tabular}{lrr}
  \hline
  Method                                           & PSNR                      & Params                     \\ \hline
  SRN ~\cite{tao2018scale}    & 35.10                     & 10.25M                     \\
  NAFNet ~ \cite{Nafnet}      & 35.48 & 67.89M \\
  Restormer~ \cite{restormer} & 35.50                     & 26.13M                     \\
  HINet ~  \cite{chen2021hinet}                 & 35.58                     & 88.67M                     \\
  BHA+ ~\cite{pan2019bringing}                                            & 36.62 & 0.51M  \\
  SRN+ \cite{tao2018scale}  & 36.87                     & 10.43M                     \\
  HINet+~\cite{chen2021hinet}            & 37.68 & 88.85M \\
  EFNet~\cite{EFNet}         & 38.12                     & 8.47M                      \\ \hline
  DLEFNet (ours)                                   & 38.40                     & 12M                        \\ \hline
  \end{tabular}
  \end{table}

\section{Conclusions}
Event cameras differ from RGB cameras in that they produce asynchronous data sequences by capturing only changes in the scene. This unique property makes it challenging to utilize existing computer vision algorithms with event data. Recent state-of-the-art solutions generate event frames by dividing the exposure time, assuming it remains constant between frames, into a fixed number of smaller time intervals. However, this approach is unrealistic as modern cameras automatically adjust exposure time based on scene information obtained from multiple sensors. Additionally, fixing the number of frames results in large time interval bins to compensate for sparse event data, causing reduced temporal resolution in scenarios with fast-moving objects or longer exposure times. To address these issues, the Deformable Convolutions and LSTM-based Flexible Event Frame Fusion Network (DLEFNet) has been proposed, demonstrating superior performance compared to existing deblurring solutions on both synthetic and real-world datasets such as GoPro and REBlur.

\bibliographystyle{splncs04}
\bibliography{biblio}
\end{document}